\documentclass[letterpaper, 10 pt, conference]{ieeeconf}  

\usepackage{color}
\usepackage{graphicx}
\usepackage{threeparttable}
\usepackage{stfloats}
\usepackage{amsfonts,amssymb}
\usepackage{verbatim}
\usepackage{amsmath}
\usepackage{balance}

\makeatletter
\newcommand{\rmnum}[1]{\romannumeral #1}
\newcommand{\Rmnum}[1]{\expandafter\@slowromancap\romannumeral #1@}
\makeatother

\IEEEoverridecommandlockouts                              

\title{\LARGE \bf
A Novel Multi-layer Framework for Tiny Obstacle Discovery
}
\author{Feng Xue$^{1}$, Anlong Ming$^{1}$, Menghan Zhou$^{1}$ and Yu Zhou$^{2,*}$
	\thanks{This work was supported by the National Natural Science Foundation of China (Nos. 61703049, 61672336, 61773295, 61876022, 61872047) }
	\thanks{*Corresponding Author}
	\thanks{$^{1}$Beijing University of Posts and Telecommunications, Beijing, China,
		{\tt\small \{xuefeng,anlongming,zhoumenghan\}@bupt.edu.cn}}%
	\thanks{$^{2}$School of Electronic Information and Communications, Huazhong University of Science and Technology, Wuhan, China,
		{\tt\small yuzhou@hust.edu.cn}}
}

\begin{document}

\maketitle
\thispagestyle{empty}
\pagestyle{empty}

\begin{abstract}
	
For tiny obstacle discovery in a monocular image, edge is a fundamental visual element.
Nevertheless, because of various reasons, e.g., noise and similar color distribution with background, it is still difficult to detect the edges of tiny obstacles at long distance.
In this paper, we propose an obstacle-aware discovery method to recover the missing contours of these obstacles, which helps to obtain obstacle proposals as much as possible.
First,
by using visual cues in monocular images,
several multi-layer regions are elaborately inferred to reveal the distances from the camera.
Second,
several novel obstacle-aware occlusion edge maps are constructed to well capture the contours of tiny obstacles,
which combines cues from each layer.
Third,
to ensure the existence of the tiny obstacle proposals,
the maps from all layers are used for proposals extraction.
Finally,
based on these proposals containing tiny obstacles,
a novel obstacle-aware regressor is proposed to generate an obstacle occupied probability map with high confidence.
The convincing experimental
results with comparisons on the \emph{Lost and Found} dataset demonstrate the effectiveness of our approach, achieving around 9.5\% improvement on the accuracy than \textit{FPHT} and \textit{PHT},
it even gets comparable performance to \textit{MergeNet}.
Moreover, our method outperforms the state-of-the-art algorithms and significantly improves the discovery ability for tiny obstacles at long distance.
	
\end{abstract}

\section{INTRODUCTION}

Autonomous vehicles on the road would be endangered by tiny obstacles, e.g., bricks, stones and lost cargos.
These obstacles (with low height of 15-30cm at long distance of 30m or more rather challenging) are hardly discovered by point clouds generated by LIDAR or stereo camera.
Moreover, the patterned ground, e.g., zebra crossing or bricks, is easily mistaken as obstacles only by appearance cues.
Hence, discovering tiny obstacles at long distance is challenging.

Most previous works \cite{LAF}\cite{PC} are based on the disparity map generated by stereo camera.
But the accuracy of disparity map is mostly harmed by optical phenomena.
Recent works \cite{MERGE}\cite{UON} define the task of obstacle discovery as segmenting the road scene into semantic labels.
They merge appearance cue and geometry cue by utilizing deep learning,
which is time-consuming.
In addition,
some methods \cite{TEN}\cite{Mao2010Preceding} utilize proposal to capture object in the image,
then build a model for classification between obstacle and non-obstacle.
However, the low perceptual ability to tiny obstacles limits such methods to detect obstacles.
Hence, the methods mentioned above are unable to extract tiny obstacles.

\begin{figure}
	\centering
	\includegraphics[width=1\linewidth]{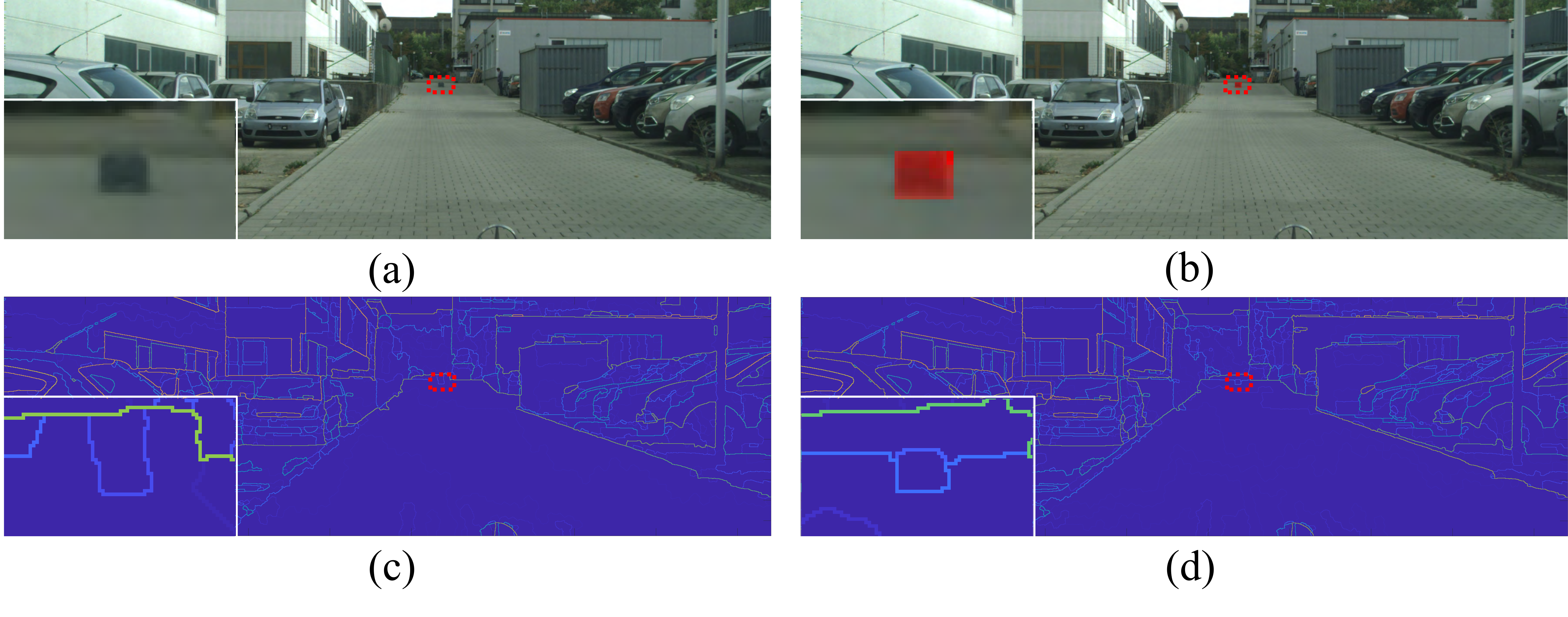}
	\caption{
		(a) is an RGB image from the Lost and Found dataset \cite{LAF}.
		(b) shows the predicted obstacle of our method which is marked in red.
		(c) denotes the occlusion edges obtained by \cite{IS}.
		(d) denotes the occlusion edges obtained by our method.
		The obstacle is enclosed in a red doted line box.
		}
	\label{fig:allv1}
\end{figure}

	Edge is an important visual element for object perception \cite{CloseEdge}.
	Furthermore,
	occlusion edge \cite{IS}\cite{Zhou2018ICPR} reveals the 3D cue of object,
	and hence captures object contours better.
	As a basic feature, it could be effectively applied in tracking \cite{Yu2016Similarity}\cite{Zhou2018ICIP}\cite{Zhou2012NIPS}, robot following\cite{Zhou2014IJPRAI}, action recognition\cite{Yu2016Human} and visual homing\cite{Ma2018Visual}, etc.
	However, in some cases like Fig.\ref{fig:allv1}(c),
	the edges of tiny objects at long distance are weak and inconsistent,
	so that the occlusion edges are insufficient acquired,
	making the proposals fail to enclose objects.
	In this paper,
	the task of obstacle discovery is based on three goals:
	1) detecting the contours of distant obstacles as completely as possible,
	2) extracting the proposals of obstacles as much as possible,
	3) ranking the obstacle proposals as high as possible.
	To achieve these goals,
	a set of obstacle-aware occlusion edge maps is generated to critically fit the contours of obstacles at various distances.
	In this generation,
	the multi-layer regions revealing the distance from camera, i.e., pseudo distance, are inferred from 2D images by visual cues.
	To enhance the perception of tiny obstacles,
	the edge cues at all pseudo distances are fused.
	Compared to the previous works \cite{IS} shown in Fig.\ref{fig:allv1}(c),
	our method fits the contours of tiny obstacles better,
	as shown in Fig.\ref{fig:allv1}(d).
	Furthermore, the proposals are extracted from the occlusion edge maps of each layer,
	ensuring the existence of tiny obstacle proposals.
	Finally, an obstacle-aware regressor based on random forest is learned to produce an obstacle occupied probability map,
	and the predicted 
	obstacles are
	shown in Fig.\ref{fig:allv1}(b).

In summary, the main contributions of our method lie in:
\begin{enumerate}
	\item A set of novel obstacle-aware occlusion edge maps is proposed to characterize obstacles,
	which fuses edge cues from each layer in a multi-layer architecture.
	These maps have a better expression for the contours of obstacles at long distance than previous works.
	
	\item By novelly combining appearance features and pseudo distance features, an obstacle-aware regressor is proposed to give high score to the obstacle proposals.
	\item
	Our method achieves remarkable performance on the \textit{Lost and Found} dataset \cite{LAF},
	outperforms the state-of-the-art algorithms and significantly improves the discovery ability for tiny obstacles at long distance.
\end{enumerate}

\section{Related work}

In general, 
there are three main categories for visual obstacle discovery:
the correlation-based methods \cite{LAF}\cite{PC},
the segmentation-based methods \cite{MERGE}\cite{UON},
and the proposal-based methods \cite{TEN}\cite{Mao2010Preceding}.
The first type compares the relative positions between 3D points in disparity map,
and classifies all points into obstacle and road.
\textit{PHT} and \textit{FPHT} \cite{LAF} apply the statistical hypothesis testing to assess the drivable area and obstacle hypotheses.
However, due to the dependence of these methods on the accuracy of disparity map,
once the disparity map is inaccurate, discovery failure would occur, especially for the tiny obstacles at long distance.
The second type segments an RGB image into several regions with different semantics.
\textit{MergeNet} \cite{MERGE} proposes a network architecture for discovering obstacles,
which makes effective use of the limited data.
However, tiny obstacles provide little information throughout the image,
hence it is hard to discover them.
The third type discovers obstacles from proposals by classification or regression.
\cite{TEN} produces plentiful proposals by \textit{Faster-RCNN} \cite{FastRCNN},
and classifies proposals by Support Vector Machines.
Methods of this type are rarely used to discover tiny obstacles,
the reason is that tiny objects do not have sufficient information.
All methods above rarely pay attention to better discover tiny obstacles at long distance.

Different from these methods mentioned above,
the focus of this paper is on the tiny obstacles at long distance.
Since our method closely depends on \textit{occlusion edge} \cite{IS} and \textit{Edge boxes} \cite{EB},
brief introductions for them are given below.

\subsection{Reviewing Edge boxes}
To model the observation of objects in an image,
\textit{Edge boxes} \cite{EB} densely searches bounding boxes in the image,
and defines the specific objectness score based on the edge map of this image.

However, some tiny obstacles have similar color distribution with road area,
hence the contours of them are incompleted and weak.
The boxes intersecting the weak edge obtain the higher score ranking.
Meanwhile, since there is no spatial constraint between pixel values in the edge probability map, 
different edge pixels of the same obstacle have completely different probabilities,
making this obstacle less like an object.
Furthermore, 
since the tiny obstacles at long distance get lower score than other larger objects,
the ability to find them greatly drop.
Hence, 
designing a method to improve the edge with closed region is necessary.

\subsection{Reviewing Occlusion Edge}

\textit{Occlusion edge} \cite{IS} aims to find the edge revealing the depth discontinuity between obstacle and background.
It takes the edges between adjacent regions of over-segmented image as inputs,
and classifies all the edges into two subsets: occlusion edges and trivial edges.
Compared to other edge cues \cite{SED} \cite{UCM},
the occlusion edge has stronger response to obstacle contour,
especially for tiny obstacles.
The reason is that the surface cue is additionally taken into account.
Hence, the occlusion edge is more proper for obstacle discovery.

However, due to the complexity of road scene,
e.g., the similar appearance with the road plane,
some tiny obstacles at long distance still hardly acquire sufficient occlusion edge cues.
The intrinsic reason is that the space distance between obstacle and camera is not taken into consideration.
Apparently, this issue brings great difficulty to discover tiny obstacles,
which would be addressed in this paper.

\section{Method}
In some cases where the tiny obstacles are located at long distance, the occlusion edge cues in \cite{IS} cannot be acquired sufficiently,
which restrict the discovery for these obstacles.
To address this issue,
a set of novel obstacle-aware occlusion edge maps is constructed to refinedly fit the contours of tiny obstacles at various distances.
Each map at multi-layer distances is used for proposal extraction,
ensuring the existence of tiny obstacle proposals.
In addition, to give relatively high scores for the tiny obstacles in proposal set,
an obstacle-aware regressor is learned by some novel features which are related to the pseudo distance.
And an obstacle occupied probability map is generated by the regressor.

\begin{figure}
	\centering
	\includegraphics[width=1\linewidth]{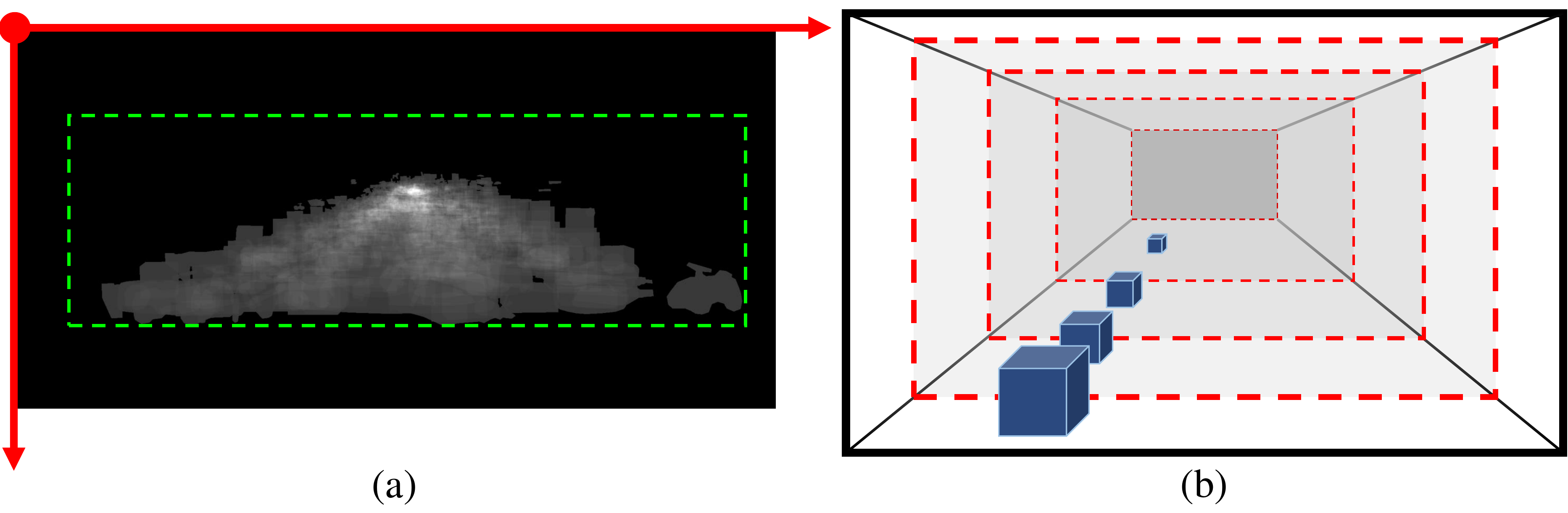}
	\caption{
		(a) the green box enclosing all the obstacles.
		(b) a toy model revealing the principle of perspective.
		For a fixed size object, 
		the farther it is in 3D space,
		the smaller it is in 2D image and farther away from the image bottom.
	}
	\label{fig:rv2}
\end{figure}

\subsection{Obstacle-aware occlusion edge map}

\begin{figure}
	\centering
	\includegraphics[width=1\linewidth]{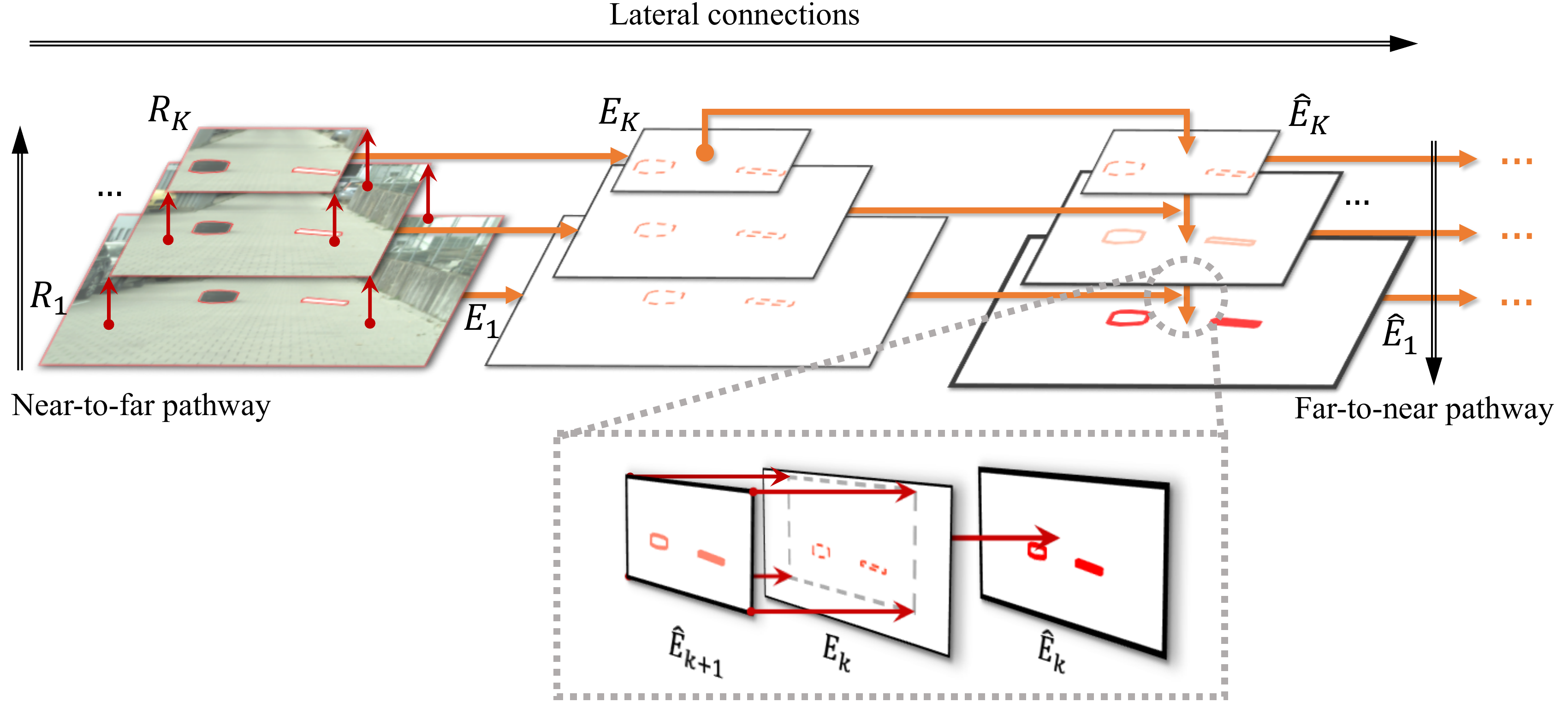}
	\caption{
		A simplified model containing the near-to-far pathway, the far-to-near pathway, and lateral connections.
	}
	\label{fig:pym}
\end{figure}

To ensure the existence of the tiny obstacle proposals,
more reliable occlusion edge maps are expected to be generated over multiple distances,
which refinedly fit the contours of tiny obstacles.
To achieve this goal,
as shown in Fig.\ref{fig:pym},
a multi-layer framework with dual paths is built.
In this framework, 
a near-to-far pathway considers various distances at which obstacles appear,
and estimates the image regions indicating different distances.
A far-to-near pathway fuses edge probability maps at different distances to enhance the edge cues of the tiny obstacles.
The lateral connections fit the contours of obstacles at different distances respectively.

{\bf Near-to-far pathway.}
To estimate multi-layer regions revealing various distances,
the principle of perspective is used to connect 3D space distance to 2D visual cues.
As shown in Fig.\ref{fig:rv2}(b),
fixing the camera,
the farther this obstacle with fixed size is in 3D space,
the smaller it is in 2D image,
meanwhile, the farther it is away from the image bottom.
Hence, two 2D properties in monocular image are employed to describe pseudo distance of an obstacle:
(\rmnum{1}) the pixel distance from an obstacle center to the image bottom,
(\rmnum{2}) the number of pixels occupied by an obstacle.
All the training obstacles are given by the Lost and Found dataset, as shown in Fig.\ref{fig:rv2}(a),
and the green doted line region ${\bf{R}}$ is used for obstacle discovery.

To be specific, given the training obstacle set $O = \{o_1,o_2,\dots,o_N\}$,
considering pseudo distance properties mentioned above,
the k-means clustering is employed to divide the whole obstacle set $O$ into $K$ subsets,
i.e., ${\bf{O}} = \{O_1,O_2,\dots,O_K\}$.
Note that each $O_k\in \bf{O}$ is a set of obstacles with similar locations and similar areas,
and the obstacles in $O_k$ are farther than that in $O_{k-1}$.
Then for an image $I$,
the region ${\bf{R}}$ is divided into sub-regions by considering the partition of $\bf{O}$,
namely, sub-region $R_1$ contains all the obstacles of $\{\bf{O}\}$,
sub-region $R_2$ contains the obstacles of $\{{\bf{O}} \setminus O_1\}$.
Following the same way,
sub-region $R_K$ contains the obstacles of $\{{\bf{O}}\setminus O_1,O_2,\dots,O_{K-1}\}$.
Intuitively, the farthest obstacle exists in sub-region $R_K$ with smallest range.
Each sub-region corresponds to a layer in the framework.
Hence, the multi-layer map is denoted as $\{R_1,R_2,...R_K\}$.

{\bf Far-to-near pathway.}
To greatly improve the edge probability of tiny obstacles,
the edge cues in distant layer would be passed to the nearby layer,
making the tiny obstacles could be clearly observed in each layer.
A set of edge probability map $\{E_1, E_2,\dots,E_K\}$, which corresponds to the layers $R_1,R_2,\dots,R_K$, is generated by the structured edge detection \cite{SED}.
	As shown in Fig.\ref{fig:pym},
	at the beginning,
	the enhanced edge probability map $\hat{E}_K$ is equal to $E_K$.
	Due to the fact that the edge pixels in $\hat{E}_K$ are also observed in the larger edge probability map $E_{K-1}$,
	there is a pixel correspondence between $\hat{E}_K$ and $E_{K-1}$.
	Hence, the values on pixels in map $\hat{E}_K$ are summed to that of the corresponding pixels in map $E_{K-1}$ to generate the enhanced map $\hat{E}_{K-1}$.
	In the same way,
	the enhanced edge probability map $\hat{E}_k$ of each layer is passed to the map $E_{k-1}$ to generate the enhanced map $\hat{E}_{k-1}$.
	In the enhanced map $\hat{E}_k$,
	since the edge cues in the long distance regions are improved by many times,
	the tiny obstacles at long distance obtain a high response.
	Hence, these obstacles are easier to be discovered.

{\bf{Lateral connections.}}
The occlusion edges are part of the atomic edges between two adjacent regions,
however, edges in $\hat{E}_k$ are unable to partition the scene into regions.
Thus, 
while retaining the edge probability gain,
the lateral connections aim to 
fit the contours of obstacles by superpixels.

\begin{figure}
	\centering
	\includegraphics[width=1\linewidth]{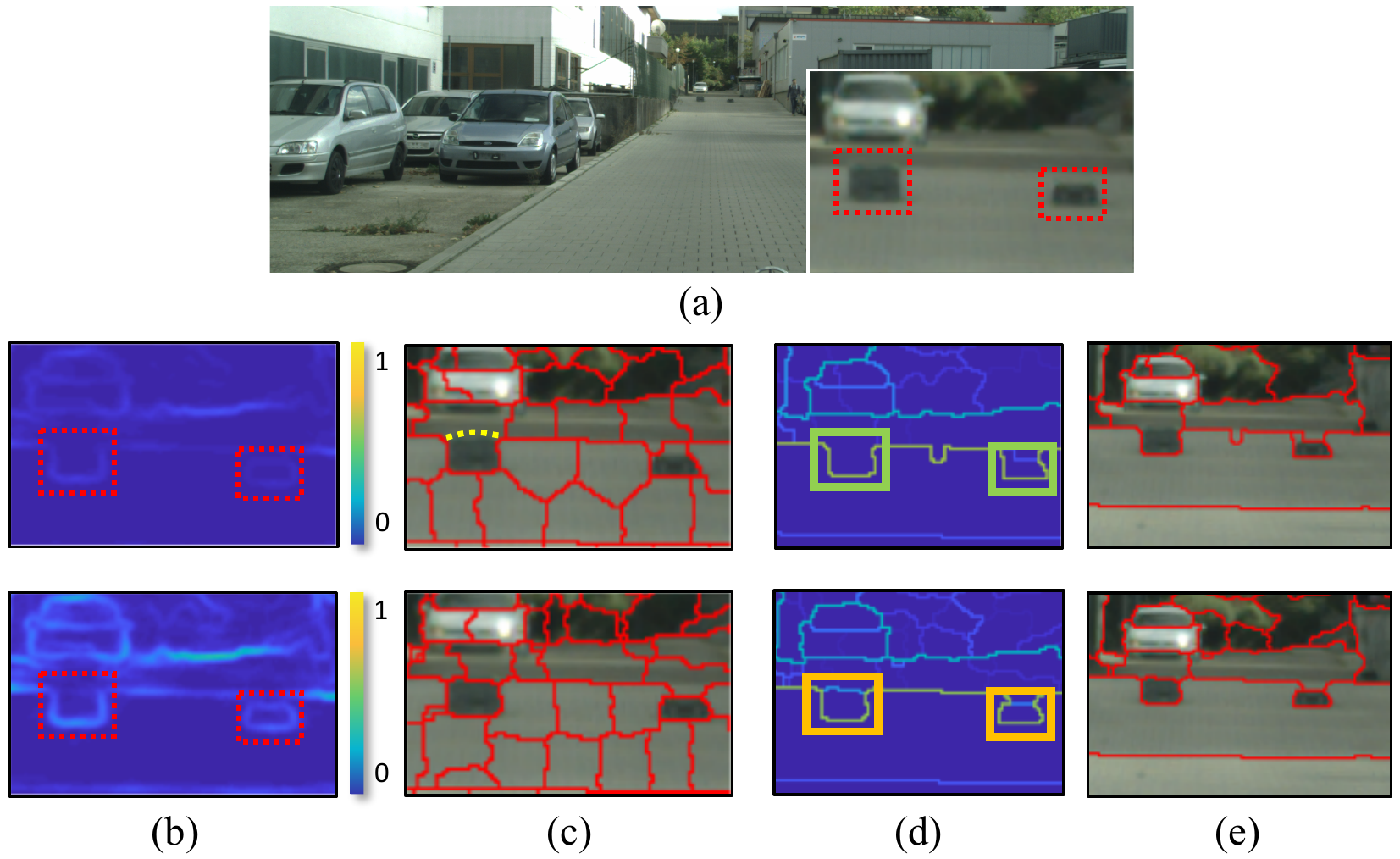}
	\caption{
		(a) the RGB image, with obstacles in the red dotted line boxes. 
		(b) the edge probability of the scene, with obstacles in the red dotted line boxes. 
		(c) the superpixels, 
		with the lost contour in the yellow dotted line.
		(d) the occlusion edges.
		(e) the occlusion edges in the red line.
		In (b)(c)(d), the top corresponds to other methods,
		the bottom corresponds to our method.
	}
	\label{fig:comp}
\end{figure}

At the beginning,
each pixel in $R_k$ is treated as a superpixel with the most detailed partition.
Then,
an iterative optimization rule proposed in \cite{SED} is used to optimize superpixels.
The enhanced edge cues are used as a critical factor to generate superpixels refereed from \cite{SED}.
The pixels inside a superpixel have similar properties.
Since the obstacles and the road area belong to different objects,
they are divided into two different superpixels.
Hence, the contours of obstacles are reflected on the atomic edges between two adjacent superpixels.
Assuming that the atomic edges set  in each layer is denoted as $S_k = \{e^k_1,e^k_2,\dots,e^k_Q\}$ corresponding to $\hat{E}_k$.
Due to the fusion of cues from various distances,
the contours of the tiny obstacles at long distance are completely fitted by the edges $e^k_q$,
as shown in the bottom of Fig.\ref{fig:comp}(c).

To capture the contours of tiny obstacles at long distance as completely as possible,
each layer generates a set of superpixels $S_k$,
and all sets $S_1,\dots,S_K$ are used for occlusion edge detection \cite{IS}.
Several cues are jointly used to express an atomic edge $e^k_q\in S_k$,
forming a feature vector $\textbf{x}^k_q\in \mathbb{R}^{V\times1}$.
The occlusion edge classifier can be trained as follows:
\begin{equation}
\min_{\mathbf{c},\textbf{b}}\sum_{n=1}^{N}(u^k_q - (\mathbf{c}^{T}\textbf{x}^k_q +\textbf{b}))^2+\gamma \mathbf{c}^T \mathbf{c}
\label{classifier}
\end{equation}
where $u^k_q$ is the class label of corresponding training edge $e_q^k \in S_k$,
$\gamma$ is a coefficient that balances the two related terms,
$\mathbf{c}$ and $\textbf{b}$ are the target classifier and bias, respectively.
The atomic edges set $S_k\subset \{S_1,S_2,...,S_K\}$ in the $k$-th layer is classified to generate $k$-th occlusion edge maps, respectively.
As shown in Fig.\ref{fig:comp}(d),
the contours of tiny obstacles below obtain higher 
scores
than 
those
of the above,
making the tiny obstacles easier to be found.

Naturally, by considering the prior cue on distance,
different layers contain obstacles at different distances.
And the tiny obstacles at long distance could be observed in scenes at all layers.
To ensure the existence of tiny obstacle 
proposals,
the occlusion edge maps in all layers are respectively used to extract proposals by the Object-Level Proposal \cite{OLP}.
Specifically, a full joint set of proposals, i.e, $B^f=B_1\cup B_2\cup\dots\cup B_K$, can be obtained, where $B_k\subset B^f$ represents the proposal set generated from $k$-th layer occlusion edge map.
However, apart from the tiny obstacles, there also exists many non-obstacles in these proposals,
e.g, the brick texture, pedestrian crossing.
It is necessary to build an obstacle-aware model for seeking the obstacles in the joint set $B^f$.

\subsection{Obstacle-aware regressor}
An obstacle-aware regressor is expected to give a high score to the real obstacle.
Due to the good generalization achieved by random forest \cite{DF},
it is suitable for the complex regression in our task.

\begin{figure}
	\centering
	\includegraphics[width=1\linewidth]{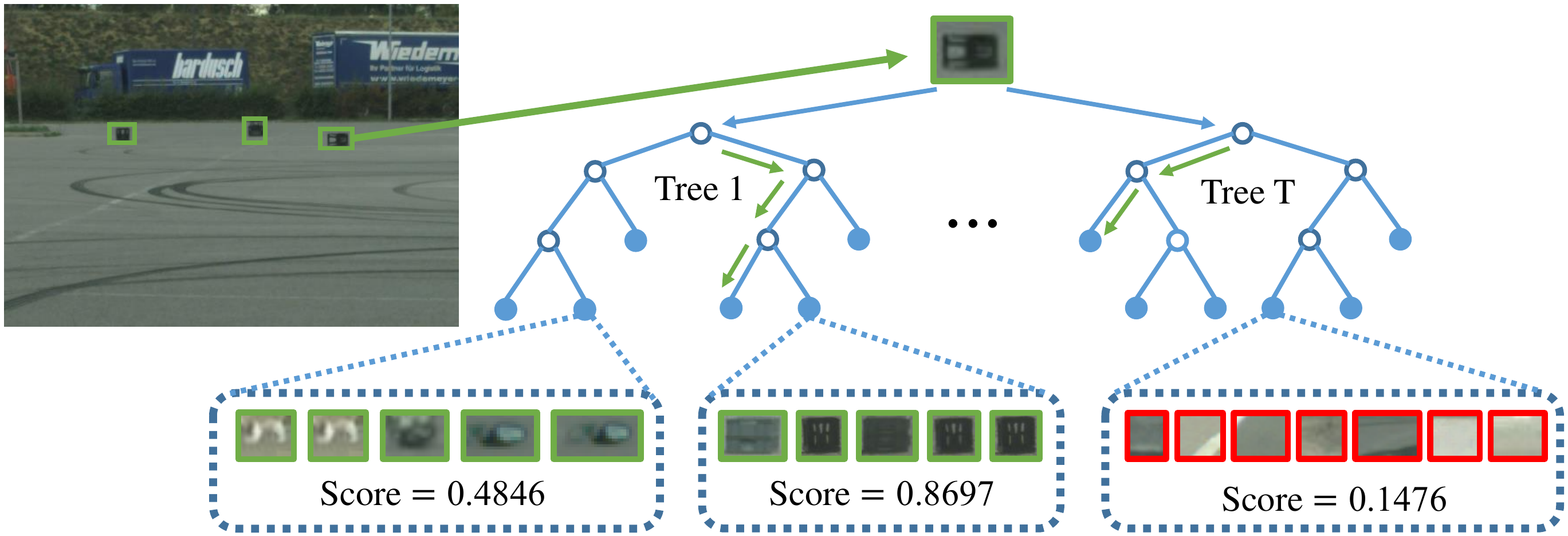}
	\caption{
		A random forest composed by a set of decision trees.
The input proposal is recursively passed to left or right child node until the leaf node is reached,
and obtains a score, i.e., an average of training proposals labels in the reach node (obstacle proposals in green, road proposals in red).
}
	\label{fig:leafv1}
\end{figure}

{\bf{Training Data Generation:}}
Training sample is one of the key elements to produce the obstacle occupied probability map.
The proposals in the initial set $B^f$ can be predefined as one of three categories:
(\rmnum{1}) road area, (\rmnum{2}) obstacle, (\rmnum{3}) non-road area. Since the harmful obstacles always lay on the road, 
only the proposals $B^r = \{b_1,b_2,\dots,b_M\}, B^r\subset B^f$, 
containing (\rmnum{1}) and (\rmnum{2}), 
are considered as the training samples.
In addition, the overlap between proposal $b_m$ and the ground truth is employed as the label.

Feature is second key element to produce the obstacle occupied probability map.
In this paper, several features are innovatively employed to characterize the proposals.
Specifically,
(\rmnum{1}) \textit{Edge and Structure}:
Edge density (ED) \cite{WO},
average, maximum, and mode of edge response,
the ratio of the mode measure the statistical information of edge;
ED measures the density of edges near the box borders.
(\rmnum{2}) \textit{Pseudo Distance}:
Following \cite{MCG}, size, position, height, width and aspect ratio of the proposal;
The combination of these features is associated to pseudo distance.
(\rmnum{3}) \textit{Objectness score}:
Following \cite{WO},
the objectness score measures the likelihood that a box contains an object.
(\rmnum{4}) \textit{Color}:
Color contrast (CC) \cite{WO} and color variance (CV) of the proposal;
CC measures the color dissimilarity of a box to its immediate surrounding area,
and CV of a box in the HSV image reflects the color dispersion inside this box.
In this work,
cosine distance between the HSV histograms is employed as the metric of CC.

Stacking all the features,
a 20 dimensional feature vector $v_{b_m}\in \mathbb{R}^{20}$ (7 for edge and structure, 6 for pseudo distance, 1 for objectness score, 6 for HSV color space) is constructed.

{\bf Obstacle-aware Regressor:} As shown in Fig.\ref{fig:leafv1},
the random forest $F = \{f_i|i=1,\dots,T\}$ consists of $T$ binary trees,
and each tree $f_i$ consists of internal nodes and leaf nodes.
The internal node classifies the proposals reaching on this node,
and passes these proposals to its left or right child node until a leaf node is reached.
And the reached leaf node stores a score which would be given to the input proposal.
Based on the generated feature vectors for training,
our regressor uses these feature vectors to regress the overlap between the obstacle and the ground truth.

\begin{figure}
	\centering
	\includegraphics[width=1\linewidth]{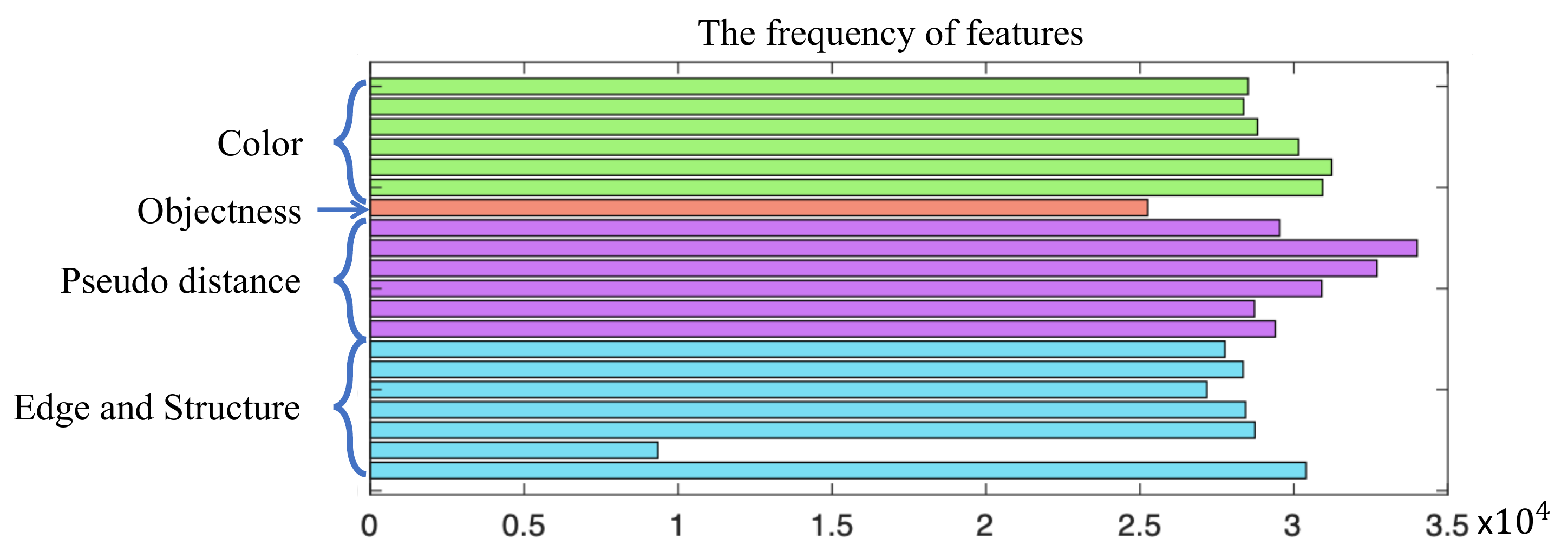}
	\caption{The frequency of the features used in the obstacle-aware regressor.}
	\label{fig:featurev1}
\end{figure}

As shown in Fig.\ref{fig:featurev1},
the frequencies of most features are similar,
which means that they have sufficient discriminability for obstacles.
The pseudo distance feature is relatively high in frequency of use,
indicating its important role for obstacle discovery.
The intrinsic reason is that the tiny obstacles at long distance have similar color with the road area,
making the color, objectness and structure properties of them different from that of the nearby obstacles.
These appearance cues are constrained by pseudo distance feature.
For the obstacles,
the combined use of all features has a higher distinguishing capability than each feature itself.

As shown in Fig.\ref{fig:leafv1},
in the forest,
the training proposals that fall inside the same leaf node have similar appearance.
It is observed that the distant lost cargos with square shape are assigned to the same leaf node.
In addition, all the obstacle proposals have convincingly higher score than the proposals containing road area,
i.e., scores of obstacles are usually higher than $0.4$,
but that of road area are lower than $0.15$.

{\bf Prediction:}
The prediction of forest $F(b_m)$ is formulated as the average of each tree output $f_i(v_{b_m})$:
\begin{equation}
F(b_m) = \frac{1}{T}\sum\nolimits_{i=1}^{T}f_i(v_{b_m})
\end{equation}
where $f_i(v_{b_m})$ denotes the output of each tree to proposal $b_m$, the score of $b_m$ is the outputs average of all trees.

{\bf Obstacle Occupied Probability Map:}
The scores of all proposals in $B^r$ are accumulated in the corresponding pixels to produce a probability map $P$.
\begin{equation}
P(pixel(\hat{p})) =
\frac{1}{\mathcal{N}^P}\sum\nolimits_{􏰔b_m\in B^r}\sum\nolimits_{\hat{p} \in b_m} F(b_m)
\end{equation}
where $pixel(\hat{p}) = (u_{\hat{p}},v_{\hat{p}})$ denotes the coordinate of pixel $\hat{p}$.
$\frac{1}{\mathcal{N}^P}$ denotes the normalization term.
If $\hat{p}$ is inside $b_m$,
$b_m$ score $F(b_m)$ is summed into $P(pixel(\hat{p}))$.
Finally, 
the tiny obstacles at long distance obtain high probabilities,
and more details are shown in the experiments.

\section{EXPERIMENTS}
\subsection{Dataset and parameter setting}
Our algorithm is validated by performing experiments on \textit{Lost and Found} dataset \cite{LAF},
i.e., the only publicly available dataset focusing on discovering the small obstacles and lost cargos on the road.
The dataset records 13 different challenging street scenarios and 37 different obstacles,
and is split into a training subset and a testing subset,
in which the obstacle types in the testing subset is more complex than that in the training subset.
For the experimental parameters,
the cluster number $K$ is set from 1 to 4 for variants comparison.
To simplify the expression, Ours@n denotes the $n$ layers variants of the multi-layer in our method.
\begin{figure}[!tp]
	\centering
	\includegraphics[width=1\linewidth]{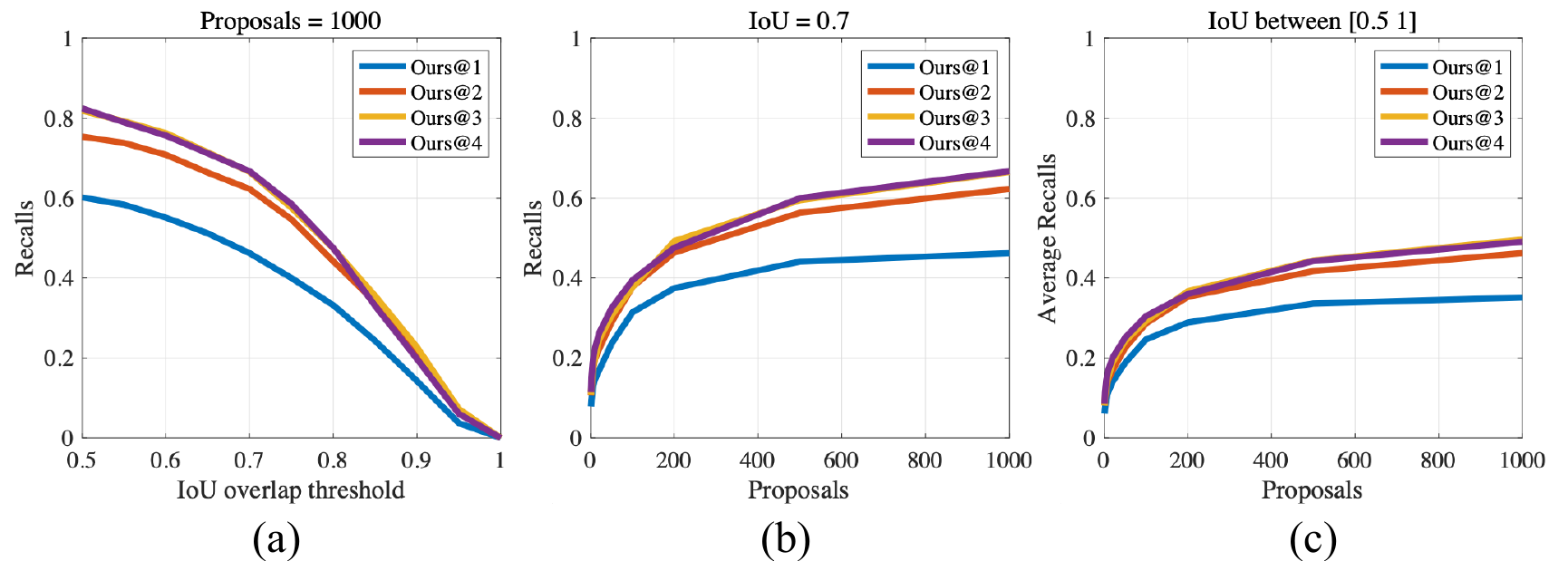}
	\caption{
		Comparison on recall rate for variants of proposed approach.
		(a) recall versus number of proposals given IoU=0.7.
		(b) recall versus IoU threshold given 1000 proposals.
		(c) average recall versus number of proposals between IoU 0.5 to 1.}
	\label{fig:propcmpselfv1}
\end{figure}
\begin{table}[!tp]
	\begin{center}
		\caption{
			Comparison on recall rate of our method variants (Taking the top 1000 proposals, the recall under different IoU. The red indicates the highest recall.)
		}
		\begin{tabular}{r|c c c c}
			\hline
			&\multicolumn{4}{c}{Variants}  \\
			IoU&\,Ours@1\,& \,Ours@2\,&\,Ours@3\,&\,Ours@4\,\\
			\hline
			0.5&0.60&0.75&0.81&\textcolor{red}{\bf 0.82}\\
			0.6 &0.55&0.70&\textcolor{red}{\bf 0.77}&0.76\\
			0.7 &0.46&0.62&0.67&\textcolor{red}{\bf 0.67}\\
			0.8 &0.33&0.44&0.48&\textcolor{red}{\bf 0.48}\\
			\hline
		\end{tabular}
		\label{table:proposal_my}
	\end{center}
\end{table}

\subsection{Evaluation metrics}
In addition, our method is evaluated on two metrics: the pixel-level metric and the instance-level metric. 

{\bf{Pixel-level Metric:}} Referring from \cite{LAF},
pixel-level Receiver-Operator-Characteristic (ROC) curve compares True-Positive-Rate (TPR) over False-Positive-Rate (FPR).
\begin{equation}
TPR = \frac{TP}{GT_{obstacle}}
\end{equation}
\begin{equation}
FPR = \frac{FP}{GT_{road}}
\end{equation}
where $TP$ denotes the correctly discovered pixel number of the obstacle,
and $FP$ denotes the number of road pixel that is incorrectly predicted as obstacle.
$GT_{obstacle}$ refers to the total pixels of the obstacle class,
and $GT_{road}$ corresponds to the road area.
In this paper,
100 thresholds from 0 to 1 are averagely taken to segment the obstacle occupied probability map,
the pixels over the threshold are labeled as obstacle.

{\bf{Instance-level Metric: }}
Three proposal metrics in \cite{OLP} are used to make comparisons on recall rate for obstacle.
Firstly, taking the top 1000 proposals, the IoU threshold ranges from 0.5 to 1.
Secondly, setting the IoU threshold to 0.7, the number of proposals ranges from 1 to 1000.
Thirdly, the average recall (AR) between IoU 0.5 to 1 is introduced,
ranging  the proposals number from 10 to 1000.

\begin{figure}[!tp]
	\centering
	\includegraphics[width=1\linewidth]{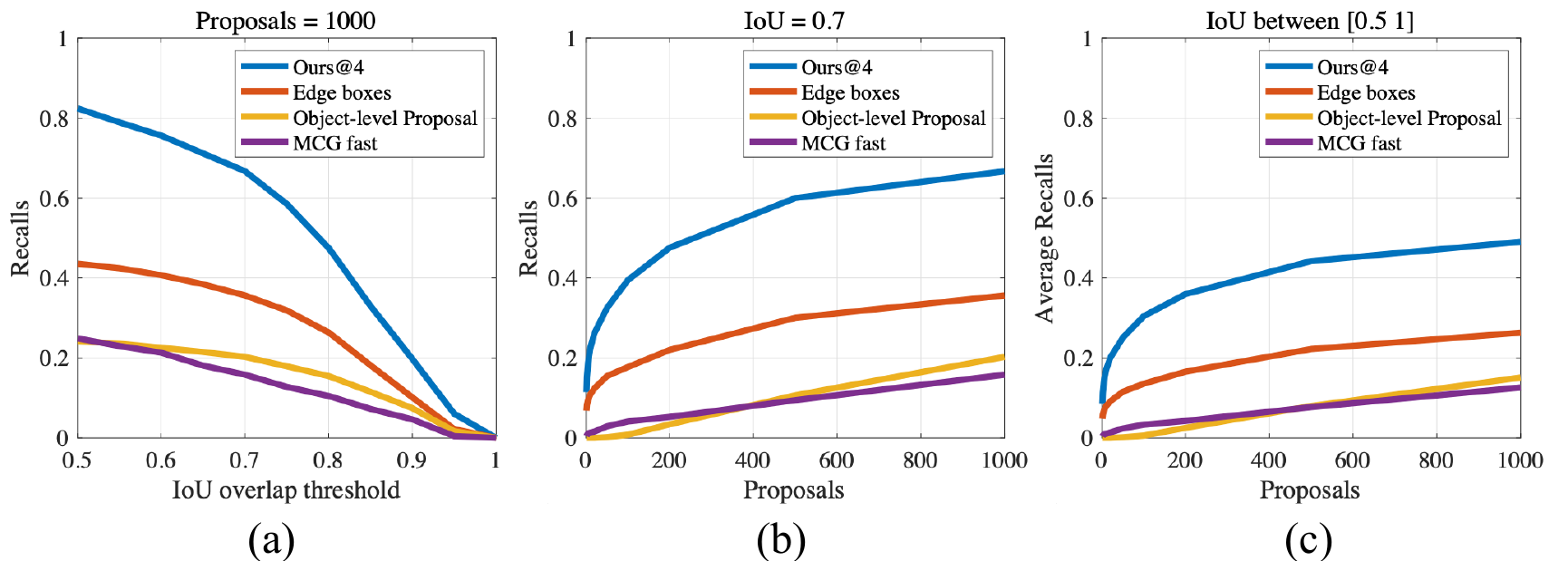}
	\caption{
		Comparison on recall rate between proposal part of our method and other proposal methods.
		(a) recall versus number of proposals given the threshold of IoU is 0.7.
		(b) recall versus IoU threshold given 1000 proposals.
		(c) average recall versus number of proposals between IoU 0.5 to 1.}
	\label{fig:propcmpothv1}
\end{figure}

\subsection{Quantitative results}
The comparison between variants of our method in pixel-level metric is shown in
Table \ref{table:roc_my}.
Note that Ours@1 has insufficient proposals when FPR is larger than 1.5\%.
Ours@4 performs favorably against the other variants,
which achieves an accuracy of 85\% when FPR is 2.1\%.
As for instance-level metric,
the comparison between variants of our method is shown in Fig.\ref{fig:propcmpselfv1} and Table \ref{table:proposal_my}.
It is observed that both Ours@3 and Ours@4 provide the best results in all experiments.
And there is a significant gap between Ours@1 and other variants.
The reason is that Ours@1 and Ours@2 hardly discover the tiny obstacles at long distance,
while Ours@3 and Ours@4 apply the layers revealing long distance to address this issue.
As the best variant, Ours@4 is used to make comparisons with the state-of-art methods below.

\begin{table}[tp]
	\caption{Comparison on ROC of our method variants (The values in red indicate the best results.)}
	\label{table:roc_my}
	\begin{center}
		\begin{tabular}{c |c c c c}
			\hline
			&\multicolumn{4}{c}{Variants} \\
			FPR&Ours@1&Ours@2 & Ours@3 & Ours@4\\
			\hline
			0.005 &  0.41  &0.57  &  0.60 &  \textcolor{red}{\bf 0.62}  \\
			0.010 & 0.52  & 0.72 & 0.74  & \textcolor{red}{\bf 0.75} \\
			0.015 &  NA  &0.78 & 0.79 & \textcolor{red}{\bf 0.80}  \\
			0.020 &  NA  &0.82 & 0.83  & \textcolor{red}{\bf 0.85}\\
			\hline
		\end{tabular}
	\end{center}
\end{table}
By utilizing the same pixel-level metric and dataset,
Table \ref{table:roctable} indicates the comparison of our method against other obstacle discovery methods.
When FPR is fixed to 2\%,
our method achieves 16\%/17\% accuracy improvement over \textit{PHT-CStix} and \textit{FPHT-CStix} \cite{LAF}, respectively.
Similarly,
when FPR is lower,
our method achieves considerable improvement in accuracy over these two methods.
\textit{MergeNet} \cite{MERGE} utilizes deep learning to discover obstacles,
and achieves an accuracy of 85\% when FPR is 2.0\%.
Although Our method is not based on deep learning,
it achieves an approximate result.

\begin{figure*}
	\centering
	\includegraphics[width=0.8\linewidth]{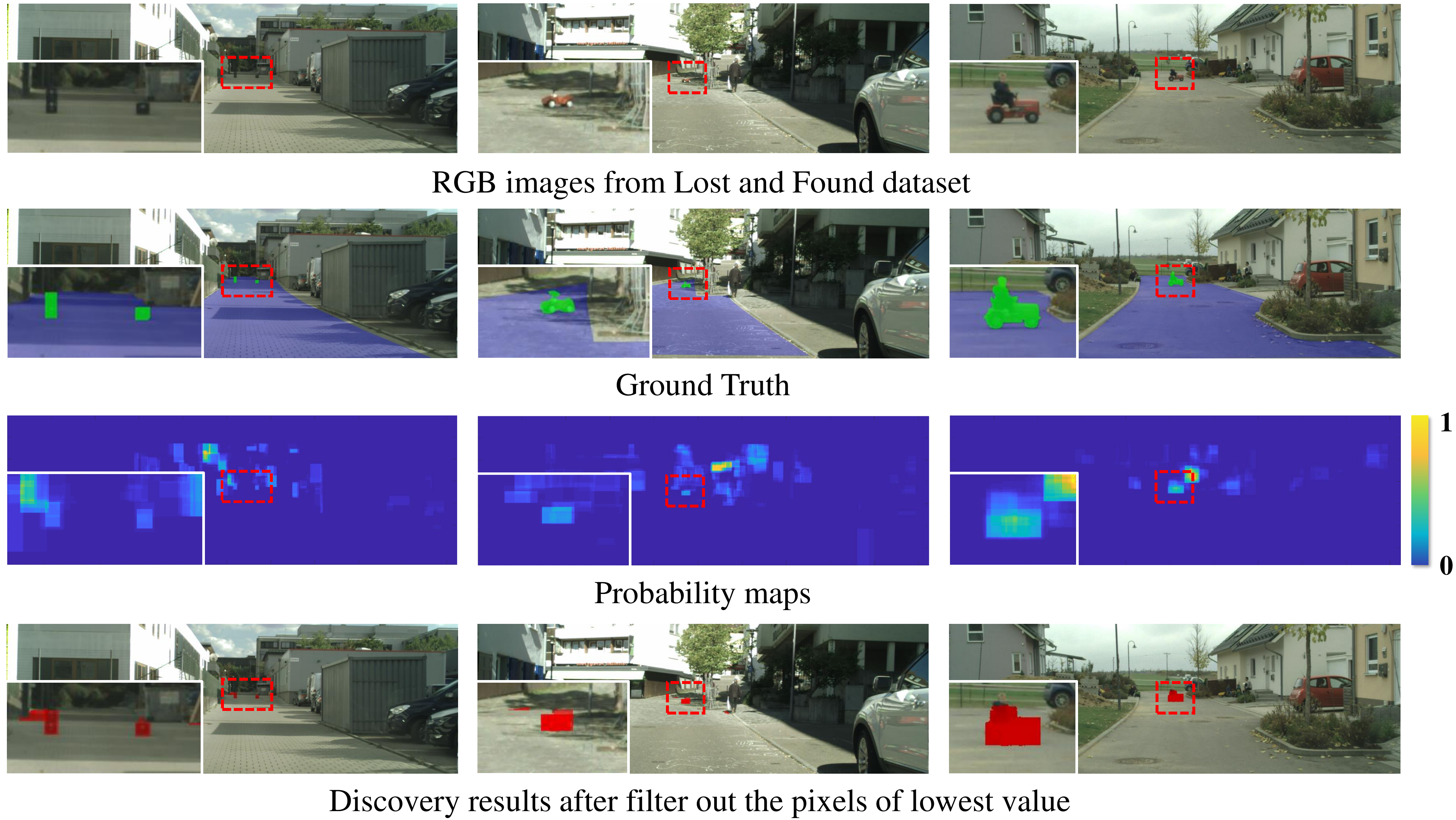}
	\caption{Qualitative results for obstacle discovery on several RGB images using our method.
		In the obstacle occupied probability maps,
		yellow colors correspond to high obstacle probability,
		and blue colors correspond to low probability.
		In the result,
		the obstacles are marked in red.
		The images could be zoomed to clearly observe these tiny obstacles.
	}
	\label{fig:allresultv1}
\end{figure*}

\begin{table}[!tp]
	\begin{center}
		\caption{
			Comparison on recall rate between our method and \cite{EB}\cite{OLP}\cite{MCG} (Taking the top 1000 proposals, the recall under different IoU and the improvement over Edge boxes. MCG(single) is the MCG variant which uses single-scale segmentation.The red values indicate the best results.)
		}
		\begin{tabular}{r|c c c c|r}
			\hline
			&\multicolumn{4}{c}{Methods}& \\
			IoU&Ours@4& Edge boxes & OLP & MCG(single) & Improvement \\
			\hline
			0.5&\textcolor{red}{\bf 0.82}&0.44&0.24&0.25&+0.39\\
			0.6 &\textcolor{red}{\bf 0.76}&0.41&0.22&0.21&+0.35\\
			0.7 &\textcolor{red}{\bf 0.67}&0.36&0.20&0.16&+0.31\\
			0.8 &\textcolor{red}{\bf 0.48}&0.26&0.16&0.10&+0.21\\
			\hline
		\end{tabular}
		\label{table:proposal}
	\end{center}
\end{table}

Another quantitative result is shown in Fig.\ref{fig:propcmpothv1} and Table \ref{table:proposal},
which compares between our method and existing proposal extraction methods on instance-level recall rate.
In Fig.\ref{fig:propcmpothv1}(a),
given 1000 proposals,
when the threshold of IoU overlap between proposals and ground truth is fixed to 0.5,
our method achieves obvious improvement in recall rate over \textit{Edge boxes}.
Meanwhile, our recall rate is also higher than that of MCG\cite{MCG} and OLP\cite{OLP}.
In Fig.\ref{fig:propcmpothv1}(b),
when the threshold of IoU overlap is fixed to 0.7,
our method always obtains the highest recall rate for different number of proposals.
In Fig.\ref{fig:propcmpothv1}(c),
for average recall (AR) versus number of proposals,
our method is also the best.
Two reasons lead to this result.
Firstly, the objectness scoring functions in these methods are too simple to express tiny obstacles on the road.
Secondly, the weak cues of tiny obstacles at long distance lead to a lower likelihood of discovering them.

\begin{table}[!tp]
		\begin{center}
			\caption{Comparison on ROC between our method and \cite{LAF} \cite{MERGE} (Red values indicate the best results.)}
			\begin{tabular}{c |c c c c}
				\hline
				&\multicolumn{4}{c}{Methods} \\
				FPR&Ours@4&PHT-CStix\cite{LAF} & FPHT-CStix\cite{LAF} & MergeNet\cite{MERGE}\\
				\hline
				0.005 &  \textcolor{red}{\bf 0.63}  &0.61  &  0.62 & NA \\
				0.010 & \textcolor{red}{\bf 0.75}  & 0.66 & 0.66  & NA \\
				0.015 &  \textcolor{red}{\bf 0.80}  &0.68 & 0.67 & NA \\
				0.020 &  \textcolor{red}{\bf 0.85}  &0.69 & 0.68  & \textcolor{red}{\bf 0.85}\\
				\hline
			\end{tabular}
			\label{table:roctable}
		\end{center}
\end{table}

\subsection{Qualitative results}
Fig. \ref{fig:allresultv1} depicts qualitative results of our methods on three challenging scenarios from the testing subset.
The left column shows cargos discarded in the shadow of buildings.
The middle column shows a bobby car parked on a chalk-marked street.
The rightmost column shows a baby on the bobby car.
Each obstacle in these scenario is very far from the camera.
In the left column,
our method completely detects distant obstacles while maintaining a low false positive in the shadow area.
In the middle column,
our method avoids detecting the ground textures and tree shades as obstacles.
In the rightmost column,
the irregular shape obstacle is not contained in the training set.
All the obstacles are successfully discovered.
Furthermore, the probability map fits the shape of obstacles by accumulating plentiful proposals.

\section{CONCLUSIONS}
In this paper, a novel obstacle discovery method is introduced.
This method proposes a multi-layer framework to produce a set of novel obstacle-aware occlusion edge maps,
which utilizes the pseudo distance.
Proposals are extracted from the occlusion edge maps of all layers,
which is able to enclose tiny obstacles as much as possible.
In addition, an obstacle-aware regressor, which fuses the pseudo distance features, is built to find obstacle proposals.
Extensive experiments validate the effectiveness of the proposed method.


\balance
\bibliographystyle{IEEEtran}
\bibliography{ICRA_OD_submitted}

\end{document}